\documentclass[10pt,twocolumn,letterpaper]{article}

\usepackage{cvpr}
\usepackage{times}
\usepackage{epsfig}
\usepackage{graphicx}
\usepackage{amsmath}
\usepackage{amssymb}

\usepackage{multirow}
\usepackage{tabularx}
\usepackage[breaklinks=true,bookmarks=false]{hyperref}

\cvprfinalcopy 

\def\cvprPaperID{226} 

\ifcvprfinal\fi
\begin{document}

\title{GS3D: An Efficient 3D Object Detection Framework for Autonomous Driving}

\author{
Buyu Li$^{1}$ \quad Wanli Ouyang$^{3}$ \quad Lu Sheng$^{1,4}$ \quad Xingyu Zeng$^{2}$ \quad Xiaogang Wang$^{1,2}$\\
$^{1}$The Chinese University of Hong Kong \quad $^{2}$SenseTime Research\\
$^{3}$The University of Sydney \quad $^{4}$Beihang University \\
{\tt\small \{byli, lsheng, xgwang\}@ee.cuhk.edu.hk, wanli.ouyang@sydney.edu.au, zengxingyu@sensetime.com}
}

\maketitle

\begin{abstract}
We present an efficient 3D object detection framework based on a single RGB image in the scenario of autonomous driving. Our efforts are put on extracting the underlying 3D information in a 2D image and determining the accurate 3D bounding box of the object without point cloud or stereo data. Leveraging the off-the-shelf 2D object detector, we propose an artful approach to efficiently obtain a coarse cuboid for each predicted 2D box. The coarse cuboid has enough accuracy to guide us to determine the 3D box of the object by refinement. In contrast to previous state-of-the-art methods that only use the features extracted from the 2D bounding box for box refinement, we explore the 3D structure information of the object by employing the visual features of visible surfaces. The new features from surfaces are utilized to eliminate the problem of representation ambiguity brought by only using a 2D bounding box. Moreover, we investigate different methods of 3D box refinement and discover that a classification formulation with quality aware loss has much better performance than regression. Evaluated on the KITTI benchmark, our approach outperforms current state-of-the-art methods for single RGB image based 3D object detection.
\end{abstract}

\section{Introduction}

3D object detection is one of the key components of autonomous driving. It has drawn increasing attention in the recent computer vision community.
With 3D LIDAR laser scanners, discrete 3D location data of objects in the form of point cloud can be fetched, but the equipment is quite expensive. On the contrary, on-board color cameras are cheaper and more flexible for most vehicles, whereas they can only provide 2D photos.
Thus 3D object detection with a single RGB camera becomes important as well as challenging for economical autonomous driving systems.
This paper focuses on detecting complete 3D object content using only monocular RGB image.

This paper proposes an efficient framework based on 3D guidance and using the surface feature for refinement (GS3D) to detect complete 3D object content using only monocular RGB image.

\begin{figure}[!ht]
\centering
\includegraphics[width=\linewidth]{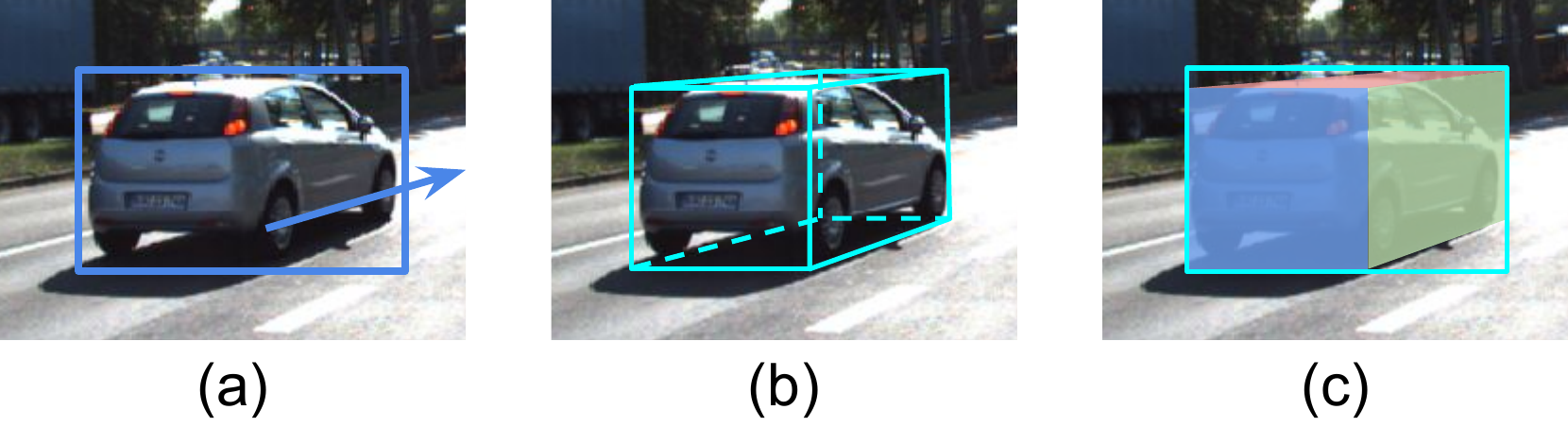}
\caption{The key idea of our method: (a) We first predict reliable 2D box and its observation orientation. (b) Based on the predicted 2D information, we utilize artful techniques to efficiently determine a basic cuboid for the corresponding object, called guidance. (c) Features extracted from the visible surfaces of projected guidance as well as the tight 2D bounding box of it will be utilized by our model to perform accurate refinement with classification formulation and quality-aware loss.}
\label{fig:key}
\end{figure}

Typical single image 3D detection methods, e.g. Mono3d~\cite{mono3d}, adopt the framework of traditional 2D detection, where exhaustive sliding windows in 3D space are utilized as proposals and the task is to select those covering the objects well. The problem is that the 3D space is much larger than the 2D space, which costs much more computation and is not necessary.

Our first observation is that a 3D coarse structure can be recovered from 2D detection and prior knowledge on the scene. 
Since state-of-the-art 2D object detection methods can provide 2D bounding boxes with quite a high accuracy, proper utilization of them can significantly reduce the search space, which is already applied in several point cloud based methods \cite{fpointnet,avod}.
Furthermore, with prior knowledge of the auto-driving scenario (e.g. the projection matrix), we can even obtain an approximate 3D bounding box (cuboid) for the object in the 2D box despite the lack of point cloud.
Inspired by this, we design an algorithm to efficiently determine a basic cuboid for the predicted object by a 2D detector. 
Although coarse, the basic cuboid has acceptable accuracy and can guide us to determine the 3D setting, size (height, width, length) and orientation of the object.
Thus the basic coarse cuboid is called \textit{Guidance} by us.

\begin{figure}[!t]
\centering
\includegraphics[width=\linewidth]{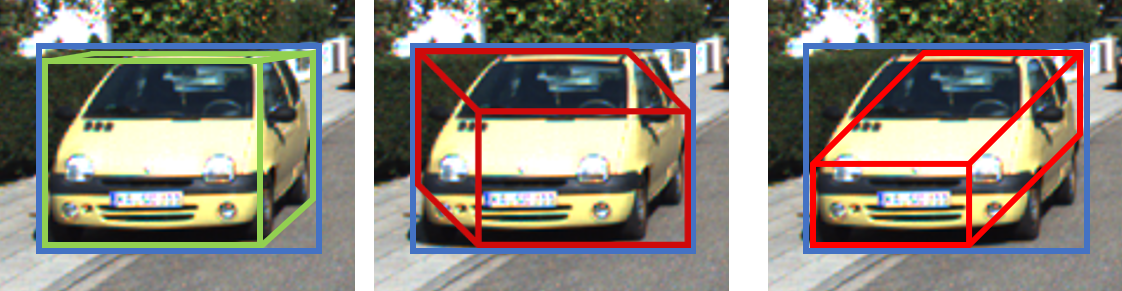}
\caption{An example of the feature representation ambiguity caused by only using 2D bounding box. The 3D boxes vary largely from each other and only the left one is correct, but their corresponding 2D bounding box are exactly the same.}
\label{fig:ambiguity}
\end{figure}

As our second observation, the underlying 3D information can be utilized by investigating the visible surfaces of the 3D box.
Based on the guidance, a further classification for eliminating false positives and appropriate refinement for better localization are necessary in order to achieve high accuracy. 
However, the information missing when using only the 2D bounding box for feature extraction brings a problem of representation ambiguity.
As shown in Fig.\ref{fig:ambiguity}, different 3D boxes varying largely from each other can just have the same corresponding 2D bounding box. Therefore the model will take the same feature as input, but the classifier is expected to predict different confidences for them (high confidence for the left one and low confidences for the others in Fig.\ref{fig:ambiguity}), which is conflict.
And the residual ($\Delta x, \Delta y$ and etc.) prediction is also difficult. From only the 2D bounding box, the model can hardly know what the original parameters (of the guidance) are, but it aims to predict the residual based on them. So training is quite ineffective.
To handle this problem, we explore the underlying 3D information in the 2D image and propose a new approach that employs features parsed from visible surfaces of the projection of the 3D box. As shown in Figure.\ref{fig:key} (c), features in the visible surfaces are extracted respectively and then incorporated, so that structural information is utilized to distinguish different forms of 3D boxes.

For 3D box refinement, we reformulate the conventional regression form into a classification form, and a quality-aware loss is designed for it, which significantly improves the performance.

Our main contributions are as follows:
\begin{enumerate}
\item We propose a purely monocular data based approach to efficiently obtain a coarse basic cuboid for the object, based on reliable 2D detection results. The basic cuboid provides a reliable approximation of the location, size, and orientation of the object and works as the guidance for further refinement. 
\item We exploit the potential 3D structural information in the visible surfaces of the projected 3D box on 2D images and propose to utilize the features extracted from these surfaces to overcome the problem of feature ambiguity in previous methods when only features from the 2D box are used. With the fusion of surface features, the model achieves the better ability of judgment and the refinement accuracy is improved.
\item We design and investigate several methods for refinement. And we draw a conclusion that discrete classification based methods with quality aware loss perform much better than direct regression approaches for the task of 3D box refinement.
\end{enumerate}

We evaluate our method on the KITTI object detection benchmark \cite{kitti}. Experiments show that our method surpasses current state-of-the-art methods using only a single RGB image and is even comparable to those using stereo data. To facilitate comparison with our works, we make our results on $val_1$ and $val_2$ available\footnote{\url{https://drive.google.com/file/d/188BxA_jlhHHpxCXk3SxPBA5qkmk53PIt/view?usp=sharing}}.

\section{Related Work}

As 3D understanding of object and scene is drawing more and more attention. Early works \cite{aspect,3drecovery,zia2013detailed,hejrati2012analyzing,del2013understanding} use low-level feature or statistics analysis to tackle 3D recognition or recover tasks. While the 3D object detection task is more challenging \cite{kitti}.

3D object detection methods can be divided into 3 categories by data, i.e. point cloud, multi-view images (video or stereo data) and monocular image.
Point cloud based methods, e.g. \cite{mv3d,fpointnet,voxelnet,avod,shi2018pointrcnn}, can directly fetch the coordinates of the points on the surfaces of objects in 3D space, so they can easily achieve much higher accuracy than the methods without point cloud. Multi-view based methods, e.g. \cite{3dop}, can obtain a depth map using the disparity computed from the images of different views. Although point cloud and stereo methods have more accurate information for 3D inference, the equipment of monocular RGB camera is more convenient and much cheaper.

The works that most related to ours are those using a single RGB image for 3D object detection in autonomous driving scenes. This setting is most challenging for the lack of 3D space information. Many recent works focus on this setting because it is a fundamental problem with great impact. Mono3d\cite{mono3d} addresses this problem through the usage of 3D sliding windows. It exhaustively samples 3D proposals from several predefined 3D regions where the objects may appear. Then it utilizes complex features of segmentation, shape, context, and location to filter out the impossible proposals and finally select the best candidates by a classifier. The complexity of Mono3d brings a serious problem of inefficiency. Whereas we design a pure projective geometry based method with a reasonable assumption, which can efficiently generate 3D candidate boxes with a much smaller number but even higher accuracy.

Since state-of-the-art 2D detectors \cite{fasterrcnn,lu2018grid,li2019gradient,liu2017recurrent,li2019zoom} can provide reliable 2D bounding boxes for objects, several works use 2D box as a prior to reduce the search region of 3D box \cite{deepmanta,deep3dbox}. \cite{deepmanta} uses a CNN to predict the parts coordinates, visibility and template similarity based on the 2D box, and match the best corresponding 3D template. While \cite{deep3dbox} first uses a CNN to predict the size and orientation based on the cropped 2D box region, and then determine the location coordinates by the constraint that the 3D box after projection should tightly fit in the 2D detection box. These methods just extract features from the 2D bounding box, which brings the problem of representation ambiguity. While we utilize surface features to eliminate the problem.

State-of-the-art monocular based methods pay more attention to the extra 3D information in order to facilitate the detection. \cite{subcnn,deepmanta,cad} try to utilize more 3D information by learning sub-categories or 3D key-points or parts in their intermediate stages. \cite{deepmanta,cad} use 2D-3D matching to determine the 3D coordinate of objects. They both need CAD models with extra labels of structure or key-points. \cite{mfusion} uses the depth information generated from disparity prediction to obtain approximate point cloud, and then use the fusion of 2D box feature and point cloud to determine the 3D box. Although only the monocular image is used in prediction, the training of the disparity model requires stereo data. In contrast to these methods, our work takes advantage of 3D structural information in the monocular image without extra data or labels.

\section{Problem Formulation}
\label{sec:form}
We adopt the 3D coordinate system from KITTI data set: the origin of the coordinate is on the camera center; $x$ axis points to right on the 2D image plane; $y$ axis points down; and $z$ axis points to the inner direction orthogonal to the image plane and stands for depth. 
3D bounding box is represented as $B = (w, h, l, x, y, z, \theta, \phi, \psi)$. Here $w, h, l$ are the size of the box (width, height, and length respectively) and $x, y, z$ are the coordinates of the \textbf{bottom} center, which is following the KITTI annotation. The size and center coordinate are measured in meter. $\theta, \phi, \psi$ are the rotation around $y$ axis, $x$ axis and $z$ axis respectively. Since our target objects are all on the ground, we only consider the $\theta$ rotation as all previous works do. 
2D bounding box is noted with a specified mark, i.e. $B^{2d} = (x^{2d}, y^{2d}, w^{2d}, h^{2d})$, where $(x^{2d}, y^{2d})$ is the center of box.

\section{GS3D}

\subsection{Overview}

\begin{figure*}[!t]
\centering
\includegraphics[width=\textwidth]{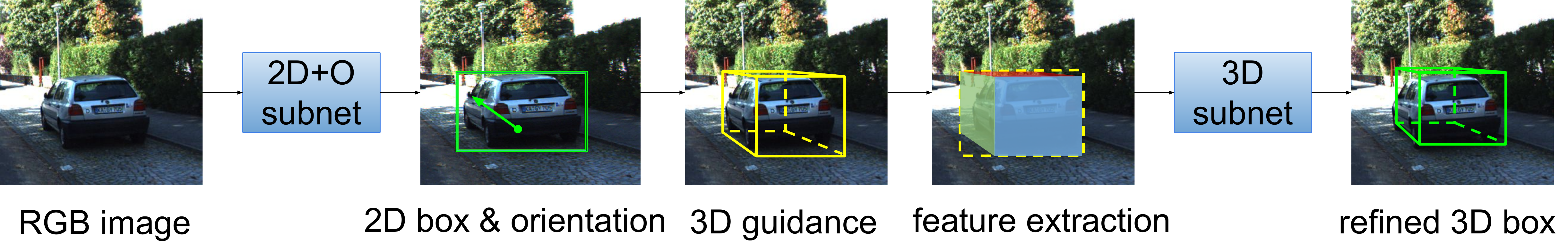}
\caption{Overview of the proposed 3D object detection paradigm. A CNN based model (2D+O subnet) is used to obtain a 2D bounding box and observation orientation of the object. The guidance is then generated by our proposed algorithm using the obtained 2D box and orientation with the projection matrix. And features extracted from visible surfaces as well as the 2D bounding box of the projected guidance are utilized by the refinement model (3D subnet). Instead of direct regression, the refinement model adopts classification formulation with the quality-aware loss for a more accurate result.}
\label{fig:framework}
\end{figure*}

Fig.~\ref{fig:framework} shows an overview of the proposed framework. 
This framework takes a single RGB image as input and consists of the following steps: 1) A CNN based detector is leveraged to obtain reliable 2D bounding boxes and observation orientations of objects. This sub-network is referred as 2D+O subnet. 2) The obtained 2D bounding box and orientation are utilized together with the prior knowledge on the driving scenario to generate a basic cuboid called guidance. 3) The guidance is projected on the image plane. Features are extracted from its 2D bounding box and visible surfaces. These features are fused as the distinguishable structural information for eliminating feature ambiguity. 4) The fused features are used by another CNN called 3D subnet to refine the guidance. The 3D detection is considered as a classification problem and quality aware classification loss is used for learning the classifiers and the CNN features.

\subsection{2D Detection and Orientation Prediction}

\begin{figure}[!h]
\centering
\includegraphics[width=0.7\linewidth]{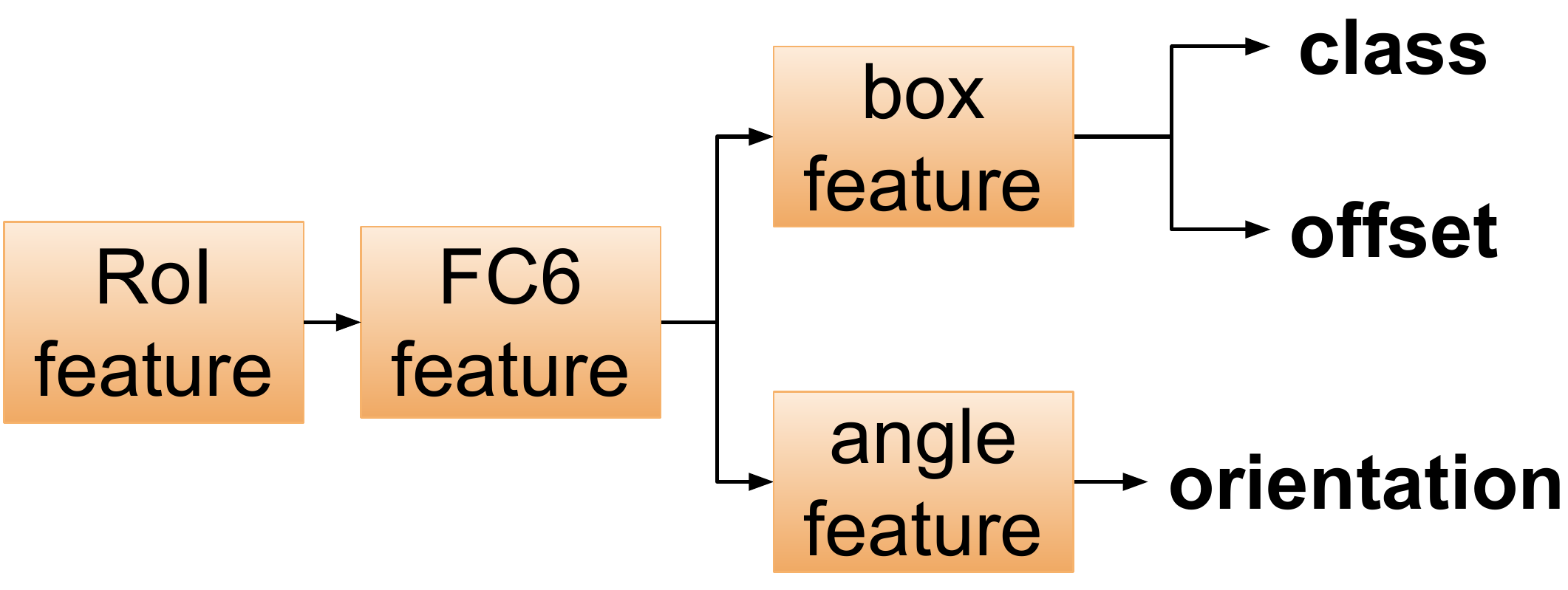}
\caption{Details of the head of 2D+O subnet. All line connections represent fully connected layers here.}
\label{fig:net_2d}
\end{figure}

For 2D detection, we modify the faster R-CNN framework by adding a new branch of orientation prediction. 
The details is illustrated in Fig.\ref{fig:net_2d}.

\begin{figure}[!h]
\centering
\includegraphics[width=0.6\linewidth]{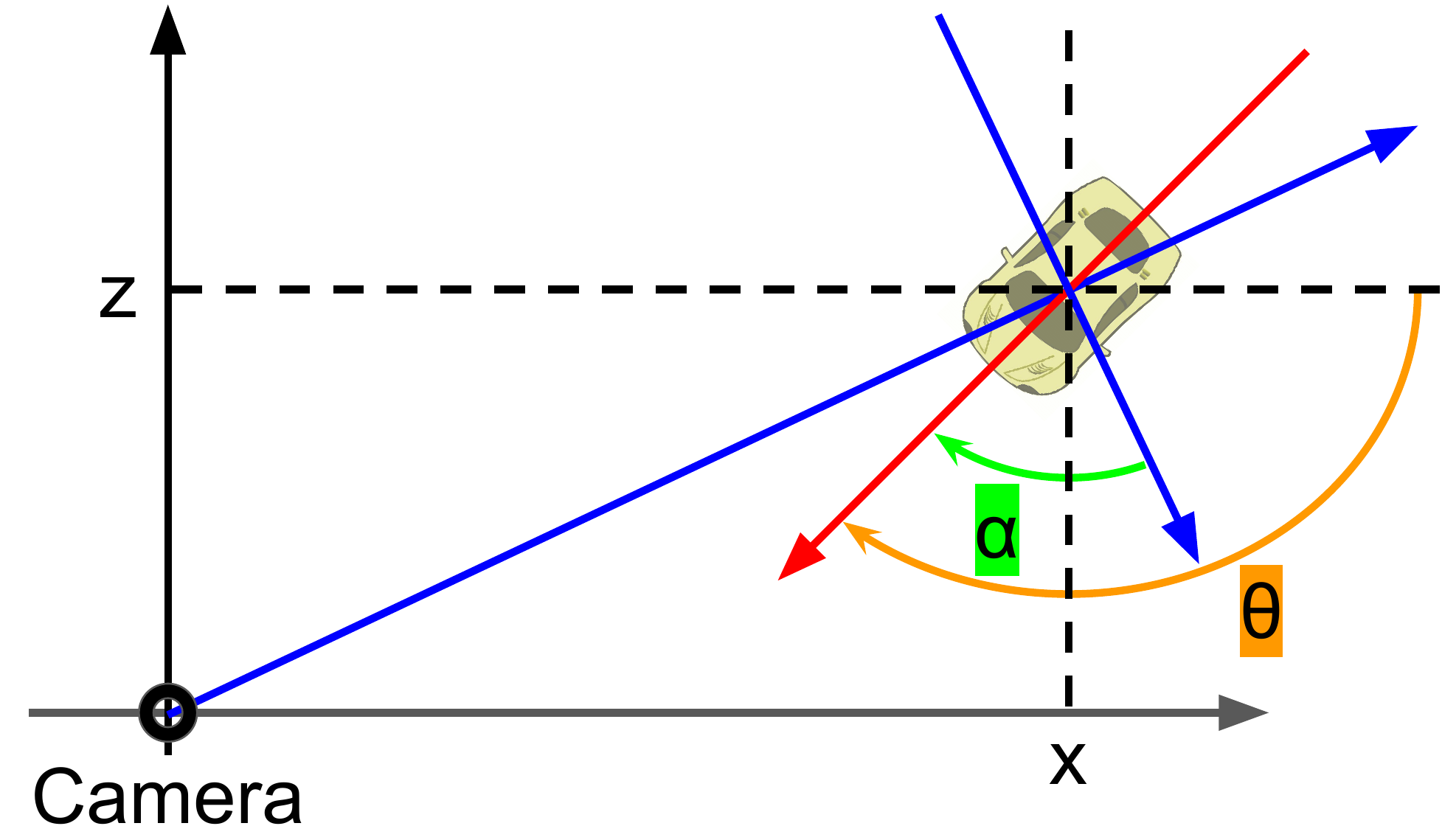}
\caption{Top view of observation angle $\alpha$ and global rotation angle $\theta$. The blue arrows represent the observation axes and the red arrow indicates the heading of the car. Since it is a right-handed coordinate system, the positive direction of rotation is clockwise.}
\label{fig:alpha2ry}
\end{figure}

Specifically, a CNN called 2D+O subnet is used for extracting features from the image, then the region proposal net generates candidate 2D box proposals. From these proposals, ROI-pooling is used for extracting the RoI features, which are then used for classification, bounding box regression, and orientation estimation.
The orientation estimated in the 2D+O subnet is the observation angle of the object which is directly related to the appearance of the object. We denote the observation angle as $\alpha$ in order to distinguish it from the global rotation, $\theta$. Both $\alpha$ and $\theta$ are annotated in the KITTI data set and their geometry relationship is shown in Fig.~\ref{fig:alpha2ry}.  

\subsection{Guidance Generation}

Based on reliable 2D detection results, we can estimate a 3D box for each 2D bounding box. 
Specifically, our target is to obtain the guidance $B_g = (w_g, h_g, l_g, x_g, y_g, z_g, \theta_g)$, given the 2D box $B^{2d} = (x^{2d}, y^{2d}, h^{2d}, w^{2d})$, the observation angle $\alpha$ and the camera intrinsic matrix $K$.

\subsubsection{Obtaining Guidance Size $(w_g, h_g, l_g)$}
In the auto-driving scenario, the distribution of the object sizes for instances of the same category is low-variance and unimodal. Since the object class is predicted by 2D subnet, we simply use the guidance size $(\bar{w}, \bar{h}, \bar{l})$ of a certain class calculated on the training data for the guidances with the same class. So we have $(w_g, h_g, l_g) = (\bar{w}, \bar{h}, \bar{l})$, which is class dependent (class does not appear in the equation for convenient notation).

\subsubsection{Estimating Guidance Location $(x_g, y_g, z_g)$}
As formulated in Section.\ref{sec:form}, $(x_g, y_g, z_g)$ is the bottom surface center of the guidance, denoted as $C_b$. So we study the characteristic of the bottom center and propose a well-worked approaches.

Our estimation approach is based on the discovery in the auto-driving settings. The top center of the object 3D box has a stable projection on the 2D plane that is very close to the top midpoint of the 2D bounding box, and the 3D bottom center has a similar stable projection that is above and close to the 2D bounding box. This discovery can be explained by the fact that the top positions of most objects have the projection that are very close to the vanishing line of the 2D image since the camera is set on the top of the data collecting vehicle and other objects in the driving scenario have similar height to it.

With the predicted 2D box $(x^{2d}, y^{2d}, w^{2d}, h^{2d})$, where $(x^{2d}, y^{2d})$ is the box center, we have the top midpoint $M_t^{2d} = (x^{2d}, y^{2d} - h^{2d}/2)$ and bottom midpoint $M_b^{2d} = (x^{2d}, y^{2d} + h^{2d}/2)$.
Then we approximately have the homogeneous form of projected top center $C_t^{2d} = (M_t^{2d}, 1) = (x^{2d}, y^{2d} - h^{2d}/2, 1)$ and bottom center $C_b^{2d} = (M_b^{2d}, 1) - (0, \lambda h^{2d}, 0) = (x^{2d}, y^{2d} + (\frac{1}{2}-\lambda)h^{2d}, 1)$, where $\lambda$ is from the statistics on training data.
With the known camera intrinsic matrix $K$, we can obtain the normalized 3D coordinates $\tilde{C}_b = (\tilde{x}_b, \tilde{y}_b, 1)$ for the guidance bottom center $C_b$ and $\tilde{C}_t = (\tilde{x}_t, \tilde{y}_t, 1)$ for the top center $C_t$ as follows:
\begin{equation}
\label{eq:backproj}
    \tilde{C}_b = K^{-1} C_b^{2d},\  \tilde{C}_t = K^{-1} C_t^{2d}.
\end{equation}

If the depth $d$ is known, $C_b$ can be obtained by:
\begin{equation}
    C_b = d\tilde{C}_b.
\end{equation}

So our target now is to obtain $d$.
We can calculate the normalized 3D coordinate of top center $\tilde{C}_t = (\tilde{x}_t, \tilde{y}_t, 1)$ by Equation (\ref{eq:backproj}). With both the bottom center and the top center, we have the normalized height $\tilde{h} = \tilde{y}_b - \tilde{y}_t$. Since the guidance height $h_g = \bar{h}$ is already obtained, we have $d = h_g / \tilde{h}$. And finally we have $(x_g, y_g, z_g) = C_b = (d\tilde{x}_b, d\tilde{y}_b, d)$.

\subsubsection{Calculating Guidance Orientation $\theta$} 
From Fig.\ref{fig:alpha2ry} we can see that the relationship between the observed angle $\alpha$ and global rotation angle $\theta$ is 
\begin{equation}
\label{eq:alpha2ry}
    \theta = \alpha + \arctan\frac{x}{z}
\end{equation}
Since $x_g, z_g$ and $\alpha$ are available through previous estimation, we can obtain $\theta_g$ by Equation.\ref{eq:alpha2ry} now.

\subsection{Surface Feature Extraction}
\label{sec:surface}
\begin{figure}[!h]
\centering
\includegraphics[width=0.9\linewidth]{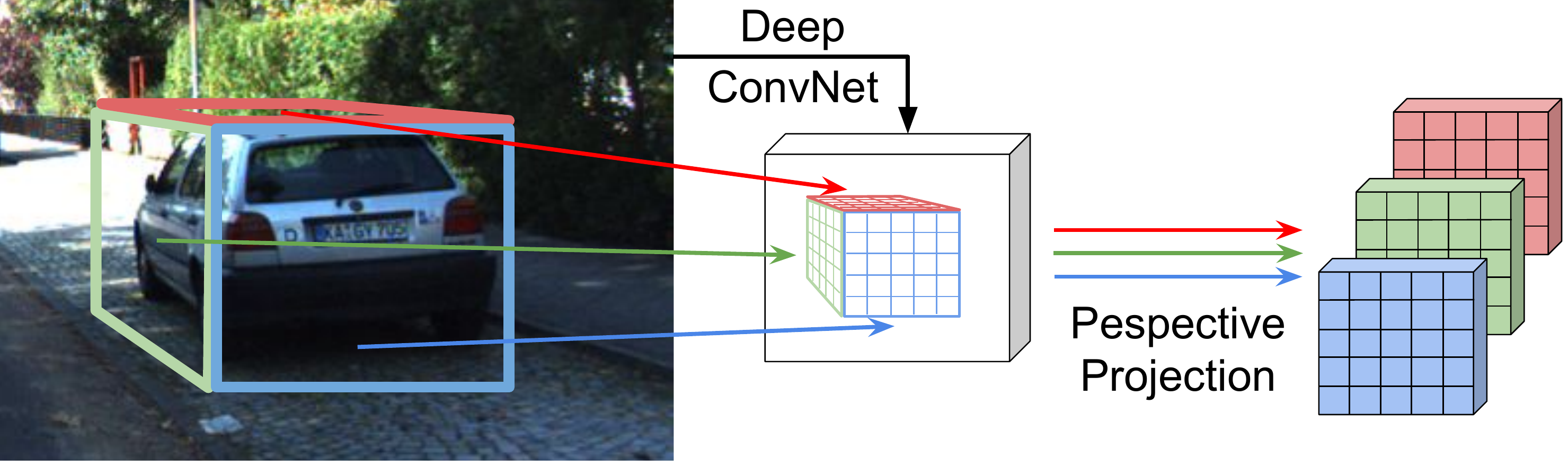}
\caption{Visualization of feature extraction from the projected surfaces of 3D box by perspective transformation.}
\label{fig:surface}
\end{figure}

We use the projected surface regions of the given 3D box (guidance) to extract 3D structure specified features for more accurate determination.
An example is illustrated in Fig.\ref{fig:surface}, the visible projected surfaces correspond to the top, left side and back of the object shown in light red, green and blue respectively. 

Since all the target objects are on the ground, the bottom surface is always invisible, we use the top surface to extract features. For the other 4 surfaces, the visibility of them can be determined by the observation orientation $\alpha$ of the object. In the KITTI coordinate system illustrated in Fig.\ref{fig:alpha2ry}, we have $\alpha \in (-\pi, \pi]$ with the right-hand direction of observer as zero angle ($\alpha=0$) and the clockwise direction as positive rotation. So when $\alpha > 0$ the front surface is visible and when $\alpha < 0$ the back surface is visible. The right side is visible when $-\frac{\pi}{2} < \alpha < \frac{\pi}{2}$, and otherwise the left side is visible.

Features in visible surface regions are warped to a regular shape (e.g. 5x5 feature map) by perspective transformation. Specifically, for a visible surface $F$, we first use the camera projection matrix to obtain the quadrilateral $F^{2d}$ in the image plane and then calculate the scaled quadrilateral $F_s^{2d}$ on the feature map according to the stride of the network. With the coordinates of the 4 corners of $F_s^{2d}$ and the target 4 corners of the 5x5 map, we can obtain the perspective transformation matrix $P$. Let X, Y represents the feature maps before and after the perspective transformation respectively. The value of the element on Y with coordinate (i,j) is computed by the following equations:
\begin{equation}
    \begin{aligned}
        Y_{i,j} &= X_{u,v} \\
    (u, v, 1) &= P^{-1} (i, j, 1)    
    \end{aligned}
\end{equation}
Usually (u,v) is not an integer coordinate and we use the 4 nearest integer coordinates with bi-linear interpolation to obtain the value $X_{u,v}$.

\begin{figure}[!h]
\centering
\includegraphics[width=\linewidth]{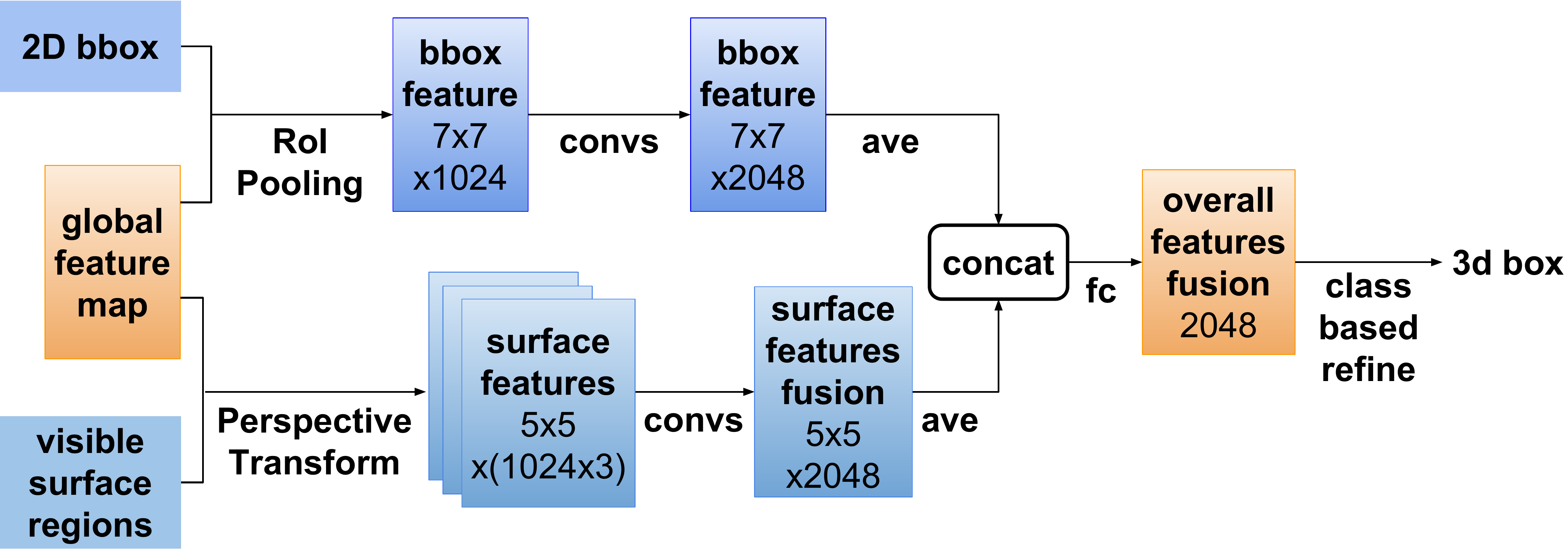}
\caption{Details of the head of 3D subnet.}
\label{fig:net_3d}
\end{figure}

The extracted features of visible surfaces are concatenated and we use convolution layers to compress the number of channels and fuse the information on different surfaces. As shown in Fig.\ref{fig:net_3d}, we also extract features from 2D bounding box to provide context information. The 2D box features are concatenated with fused surface features, and they are finally used for refinement. 

\subsection{Refinement Methods}
\label{sec:refine}

\subsubsection{Residual Regression}
With the candidate box $(w,h,l,x,y,z,\theta)$ and target ground truth $(w^*,h^*,l^*,x^*,y^*,z^*,\theta^*)$, the residuals are encoded as:
\begin{equation}
\begin{aligned}
    &\Delta x = \frac{x^* - x}{\sqrt{l^2+w^2}} , \Delta y = \frac{y^* - y}{\sqrt{l^2+w^2}} , \Delta z = \frac{z^* - z}{h}, \\ 
    &\Delta l = \log(\frac{l^*}{l}) , \Delta w = \log(\frac{w^*}{w}) , \Delta h = \log(\frac{h^*}{h}) , \\
    &\Delta \theta = \theta^* - \theta 
\end{aligned}
\end{equation}
And the commonly used method is to predict the encoded residuals by regression model.

\subsubsection{Classification Formulation}
\label{sec:class}
Regression in a large scope usually performs no better than discrete classification, so we transform the residual regression into a classification formulation for 3D box refinement.
The main idea is to divide the residual range into several intervals and classify the residual value into one interval.

Denote $\Delta d_i = d_i^{gt} - d_i^{gd}$ as the difference of the $i$th guidance and its corresponding ground-truth 3D setting descriptor $d$ where $d \in \{w,h,l,x,y,z,\theta\}$. 
The standard deviation $\sigma(d)$ of $\Delta d$ on the training data is calculated. Then we assign $(0, \pm\sigma(d), \pm2\sigma(d), ..., \pm N(d)\sigma(d))$ as the center for the intervals of descriptor $d$ and each interval has a length of $\sigma(d)$. $N(d)$ is chosen according to the range of $\Delta d$.

Since the guidance may come from a false positive 2D box, we treat the intervals as multiple binary classification problems. During training, if the 2D bounding box of the guidance cannot be matched with any ground-truth, the probability for all the intervals will be close to 0.  In this way, we can consider the guidance to be a background and reject it during inference if the confidences of all classes are very low. 

\subsubsection{Classification after Shift}
\label{shift}
Since mapping 2D regions to 3D space is an under-determined problem, we further consider starting from deviations directly in the 3D coordinate. Specifically, each class (residual interval) uses the most correlated region (the projection of guidance after corresponding residual shift) to extract individual features for itself. And all the classifiers of residual intervals can share parameters.

\subsubsection{Quality Aware Loss}
\label{sec:loss}
We expect the confidence predicted in classification to reflect the quality of the target box of corresponding class, so that the more accurate target box gets the higher score. This is important because AP (average precision) is computed by sorting the candidates with respect to their scores. 
However, the common used 0/1 label is improper for the purpose because the model is forced to predict 1 for all positive candidates regardless of their variation in quality. Inspired by loss in 2D detection \cite{iounet}, we change the 0/1 label to a quality aware form:
\begin{equation} 
\label{eq:label}
  q =\left\{
    \begin{aligned}
    & 1  & \text{if } ov > 0.75 \\
    & 0 & \text{if } ov < 0.25 \\
    & 2ov - 0.5 & \text{otherwise}
    \end{aligned}
  \right.
\end{equation}
where $ov$ is the 3D overlap between the target box and ground-truth.
And we use BCE as the loss function:
\begin{equation}
\label{eq:bce}
  L_{quality} = -[q\log(p) + (1-q)\log(1-p)].
\end{equation}

\begin{table*}[!t]
\small
\begin{center}
\begin{tabular}{| c | c | c | c@{/}c | c@{/}c | c@{/}c || c@{/}c | c@{/}c | c@{/}c |}
\hline
\multirow{2}{*}{Method} & \multirow{2}{*}{Extra} & \multirow{2}{*}{Time} & \multicolumn{6}{c||}{$\text{AP}_{3D}$ (IoU=0.5)} & \multicolumn{6}{c|}{$\text{AP}_{3D}$ (IoU=0.7)} \\
\cline{4-15}
& & & \multicolumn{2}{c|}{Easy} & \multicolumn{2}{c|}{Moderate} & \multicolumn{2}{c||}{Hard} & \multicolumn{2}{c|}{Easy} & \multicolumn{2}{c|}{Moderate} & \multicolumn{2}{c|}{Hard} \\
\hline     
Deep3DBox \cite{deep3dbox} & None & - & 27.04 & - & 20.55 & - & 15.88 & - & 5.85 & - & 4.10 & - & 3.84 & - \\
Mono3D \cite{mono3d} & Mask & 4.2 s & - & 25.19 & - & 18.20 & - & 15.52 & - & 2.53 & - & 2.31 & - & 2.31 \\
3DOP \cite{3dop} & Stereo & 3 s & - &  \textbf{46.04} & - &  \textbf{34.63} & - &  \textbf{30.09} & - & 6.55 & - & 5.07 & - & 4.10 \\
MF3D \cite{mfusion} & Stereo & - & \textbf{47.88} & 44.57 & 29.48 & 30.03 & \textbf{26.44} & 23.95 & 10.53 & 7.85 & 5.69 & 5.39  & 5.39 & 4.73 \\
\hline
Ours & None & 2.3 s & 34.72 & 33.11 & \textbf{30.06} & 27.16 & 24.78 & 23.57 & 9.12 & 8.71 & 6.71 & 6.64 & 6.31 & 6.11 \\ 
Ours (scls) & None & 2.3 s & 32.15  & 30.60 & 29.89  & 26.40 & 26.19 & 22.89 & \textbf{13.46} & \textbf{11.63} & \textbf{10.97} & \textbf{10.51} & \textbf{10.38} & \textbf{10.51} \\ 
\hline
\end{tabular}
\vspace{1.5mm}
\caption{3D detection accuracy on KITTI for car category evaluated using the metric of $\text{AP}_{3D}$. Results on the two validation sets $val_1$ / $val_2$. ``Extra'' means the extra data or label used in training. ``scls'' represents the method using shift feature for classification. }
\label{tab:3d}
\end{center}
\end{table*}

\section{Experiments}

We evaluate our framework on KITTI object detection benchmark \cite{kitti}. It consists of 7,481 training and 7,518 test images. We follow \cite{deepmanta} to use two train/val splits. Among the previous works, \cite{3dvp,deep3dbox} use $val_1$, and \cite{mono3d,3dop} use $val_2$, and \cite{deepmanta,mfusion} use them both. Our experiments are focused on the car category like most previous works do.

\subsection{Implementation Details}

\subsubsection{Network Setup:}
Both our 2D sub-net and 3D sub-net are based on the VGG-16 \cite{vgg} network architecture. The 2D sub-net takes a classification model pre-trained on ImageNet data set to initialize its parameters. And the trained model of 2D sub-net is used to initialize the parameters of 3D sub-net in training.

\subsubsection{Optimization}
We use the Caffe deep learning framework \cite{caffe} for training and evaluation. During training, we upscale the image by a factor of 2, and use 4 GPUs with one image on each. We run SGD solver with a base learning rate of 0.001 for the first 30K iterations and reduce it to 0.0001 for another 10K iterations.

\subsection{Ablation Study}

\subsubsection{2D Detection and Orientation}
Since our efforts are focused on 3D detection, we spare no time for tunning the hyper-parameters (e.g. loss weight, anchor size) for best performance of the 2D model and just train the 2D subnet without bells and whistles. 
We evaluate the Average Precision (AP) and Average Orientation Similarity (AOS) of our 2D model, following the standard KITTI setup.
The results are shown and compared with other state-of-the-art works in Table.\ref{tab:2d}.
Despite Deep3DBox \cite{deep3dbox} with much higher AP, our result is better than or comparable to other works. Moreover, although Deep3DBox use better 2D box for 3D box estimation, our 3D results surpasses theirs by a large margin (Table.\ref{tab:3d}), which highlights the strength of our 3D box determination method.

\begin{table}[!ht]
\begin{center}
\footnotesize
\begin{tabular}{| c | c@{/}c | c@{/}c |}
\hline
Method & \multicolumn{2}{c|}{$\text{AP}_\text{2D}$} & \multicolumn{2}{c|}{$\text{AOS}$} \\
\hline
Mono3D \cite{mono3d} &- & 88.67  &- & 86.28  \\
3DOP \cite{3dop} &- & 88.07 &- & 85.80  \\
Deep3DBox \cite{deep3dbox} & \underline{97.20} & - & \underline{96.68} &  \\
DeepMANTA \cite{deepmanta} & 91.01 & \underline{90.89}  & 90.66 & \underline{90.66}  \\
\hline
Ours & 90.02 & 88.85 & 89.13 & 87.52 \\
\hline
\end{tabular}
\vspace{1.5mm}
\caption{Comparison of 2D detection and orientation results for car category evaluated on $val_1$ / $val_2$ of KITTI data set. Only the results under the moderate criteria, the primal metric of KITTI, are shown for convenient size of table.}
\label{tab:2d}
\end{center}
\end{table}

\subsubsection{Guidance Generation}
Based on the statistics on training data, we set $\bar{w}=1.62$, $\bar{h}=1.53$, $\bar{l}=3.89$ as the size of guidance and $\lambda = 0.07$ for the shift of the projected bottom center.

To better evaluate the accuracy of the guidance, we use the metric of Recall$_\text{loc}$ as well as Recall$_\text{3D}$. For Recall$_\text{loc}$, the Euclidean distance between box centers of candidates and ground truths is calculated, and the ground-truth box is recalled if there is an candidate whose distance from it is within a threshold. While Recall$_\text{3D}$ is similar with the criteria changed from distance to 3D overlap.

As shown in Table.\ref{tab:guidance}, we also compare our guidance recall with the proposals recall of Mono3D \cite{mono3d} for their similar roles in the 3D detection framework. The evaluation is performed on $val_2$.  more efficient than the complex method of proposal generating of Mono3D.

Note that the number of guidance is just equals to the number of 2D detected boxes, which is of the same order of magnitude as ground-truth. So the Recall$_\text{3D}$ of guidance is similar to AP$_\text{3D}$, and our refined 3D boxes can achieve an AP that surpasses the value of guidance Recall.

\begin{table}[!ht]
\setlength{\tabcolsep}{1mm}
\footnotesize
\begin{center}
\begin{tabular}{| c || c | c || c | c | c |}
\hline
\multirow{2}{*}{Method} & \multicolumn{2}{c||}{Recall$_\text{loc}$} & \multicolumn{3}{c|}{Recall$_{\text{3D}}$@IoU=0.5} \\
\cline{2-6}
& thr=2m & thr=1m & Easy & Moderate & Hard \\
\hline
Mono3D \cite{mono3d} & 79.10 & 70.24 & 29.55 & 27.72 & 27.23 \\
Ours & \textbf{89.80} & \textbf{85.78} & \textbf{35.52} & \textbf{28.74} & \textbf{25.02} \\
\hline
\end{tabular}
\vspace{1mm}
\caption{Recall$_\text{loc}$ and Recall$_\text{3D}$ of our results compared with Mono3D. The IoU threshold of Recall$_\text{3D}$ is 0.5. These are evaluated on $val_2$ set.}
\label{tab:guidance}
\end{center}
\end{table}

\subsubsection{Refinement}
The ablation study of the contribution of surface feature, classification formulation and quality aware loss are shown in Table.\ref{tab:ablation}.

We first train a baseline model using direct residual regression following previous works e.g. \cite{3dop,mfusion}. And the baseline only uses guidance region (bounding box) features pooled from the feature map of the image.

Then we adopt the network architecture in Fig.\ref{fig:net_3d} and train a surface feature aware model. With the surface feature providing 3D structurally distinguishable information, the regression accuracy is improved (seen in the line of ``+surf'').

For the classification formulated refinement, the distributions of $\Delta d$ for each dimension on the training set are analyzed as shown in Table.\ref{tab:std}. As stated in Section.\ref{sec:class}, we set the interval length for each dimension as the $\sigma_d$. And we choose $N_d = 5$ for $d \in \{w,h,l,y,\theta\}$ and $N_x = N_z = 10$, mainly according to the range over std ratio.

\begin{table}[!ht]
\setlength{\tabcolsep}{1mm}
\footnotesize
\begin{center}
\begin{tabular}{| c | c | c | c | c | c | c | c |}
\hline
Dimension & w & h & l & x & y & z & $\theta$ \\
\hline
std & 0.10 & 0.13 & 0.41 & 0.48 & 0.10 & 1.65 & 0.05 \\
\hline
\multirow{2}{*}{range} & -0.49,  & -0.44,  & -1.74,  & -10.89,  & -0.52,  & -12.78,  & -0.27,  \\
&0.40 & 0.90 & 1.27 & 6.22 & 0.69 & 27.06 & 0.31 \\
\hline
\end{tabular}
\vspace{1mm}
\caption{Distribution analysis of $\Delta d$ on training data.}
\label{tab:std}
\end{center}
\end{table}

With the parameters for classes settled, we perform experiments with the classification formulation instead of the direct regression. Comparison experiments using the features after shift for classification are also conducted. In Table.\ref{tab:ablation}, ``+cls'' and ``+scls'' represent these two methods respectively. We can see the two class formulated methods both surpass the regression method. The fixed feature based method performs better in AP@0.5, while the shift feature based one performs better in AP@0.7. 

Finally we change the 0-1 label based loss to the quality aware form introduced in Section.\ref{sec:loss}. Significant gain is achieved in both classification based methods (seen in the line ``+qua'' of Table.\ref{tab:ablation}).

\begin{table}[!ht]
\setlength{\tabcolsep}{1mm}
\footnotesize
\begin{center}
\begin{tabular}{| c || c | c | c || c | c | c |}
\hline
\multirow{2}{*}{Method} & \multicolumn{3}{c|}{AP$_\text{3D}$@IoU=0.5} 
& \multicolumn{3}{c|}{AP$_{\text{3D}}$@IoU=0.7} \\
\cline{2-7}
& Easy & Modr & Hard & Easy & Modr & Hard  \\
\hline
Baseline & 21.66 & 15.47 & 14.75 & 2.75 & 1.99 & 1.86 \\
+surf  & 25.81 & 20.41 & 17.70 & 3.75 & 2.99 & 2.86 \\
+surf +cls & 30.87 & 23.39 & 19.86 & 5.09 & 3.76 & 3.63 \\
+surf +scls & 28.57 & 18.81 & 17.63 & 7.41 & 4.51 & 4.51 \\
+surf +cls +qua & \textbf{33.11} & \textbf{27.16} & \textbf{23.57} & 8.71 & 6.64 & 6.11 \\
+surf +scls +qua & 30.60 & 26.40 & 22.89 & \textbf{11.63} & \textbf{10.51} & \textbf{10.51} \\
\hline
\end{tabular}
\vspace{1mm}
\caption{Ablation study of 3D detection results for car category on KITTI $val_2$ set. ``Modr'' means moderate here. And ``+surf'', ``+cls'', ``+scls'', ``+qua'' represent the usage of surface feature, class formulation, shift based class formulation and quality aware loss respectively.}
\label{tab:ablation}
\end{center}
\end{table}

\begin{figure*}[!ht]
\centering
\includegraphics[width=0.9\linewidth]{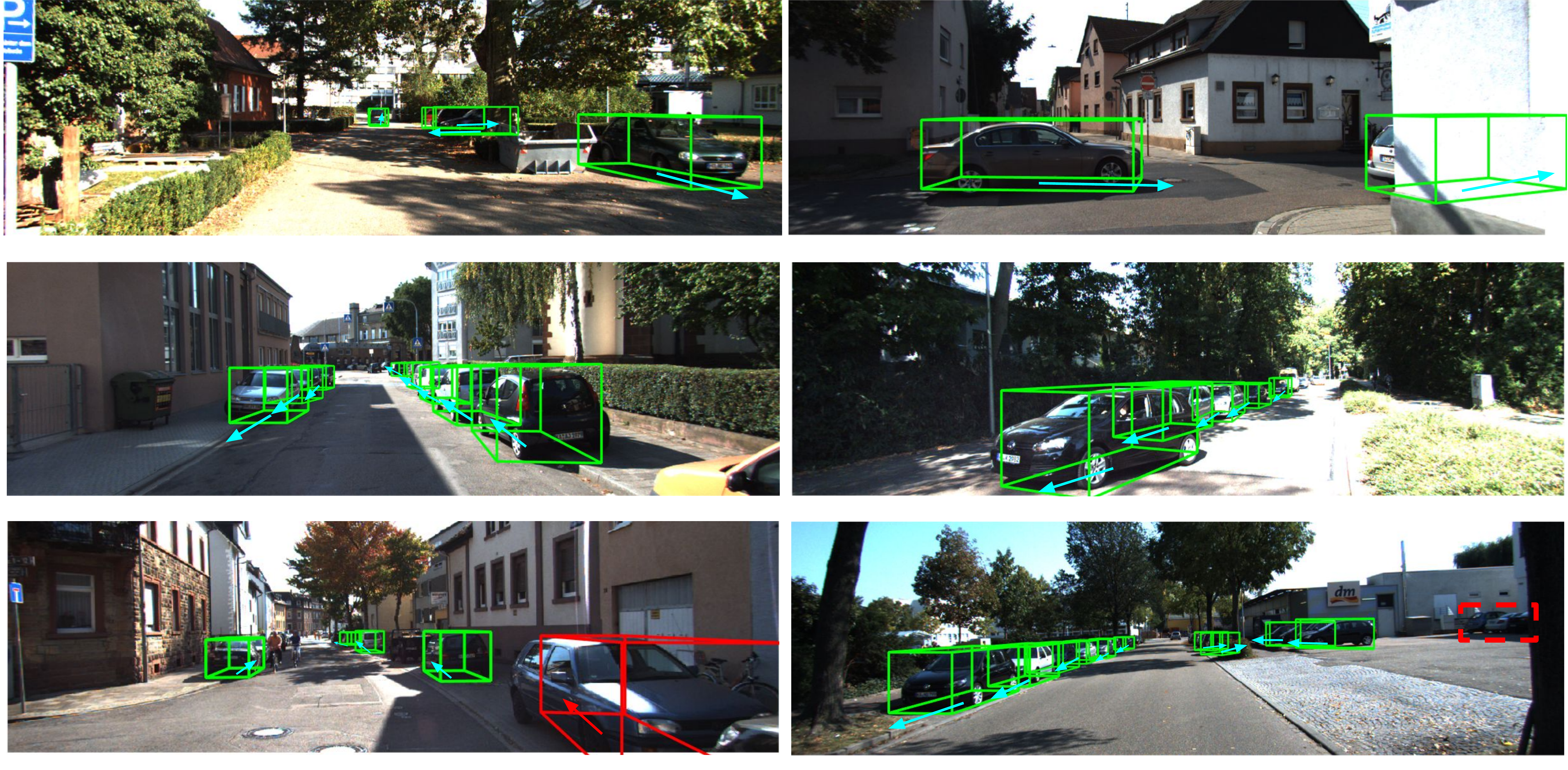}
\caption{Qualitative illustration of our 3D detection results.}
\label{fig:result}
\end{figure*}
\vspace{-3mm}

\subsection{Comparison with Other Methods}

We compare our work with state-of-the-art RGB image based 3D object detection methods:  Mono3D \cite{mono3d}, Deep3DBox \cite{deep3dbox}, DeepManta \cite{deepmanta}, MF3D \cite{mfusion} and 3DOP \cite{3dop}. 

Most of these methods requires extra data or label in addition to single RGB image and the KITTI official annotation for training. 3DOP is a stereo data based method. Mono3D need segmentation data for the mask prediction. DeepManta need 3D CAD data and vertices for their 3D model prediction. MF3D adopts the model in MonoDepth \cite{monodepth} for their disparity prediction, which is actually trained on stereo data. Whereas only Deep3DBox, as well as our work, requires no extra data or label.

\textbf{AP$_\text{3D}$:} 
The major metric for our 3D detection evaluation is the KITTI official 3D Average Precision (AP$_\text{3D}$): a detection box is considered as true positive if it has a overlap (IoU) with the ground truth box larger than the threshold IoU=0.7. We also show result comparison with IoU=0.5. As we can see in Table.\ref{tab:3d}, our method surpasses other works by a large margin in the official metric (IoU=0.7), while 3DOP has a better performance evaluated with IoU=0.5. This indicates that our method can achieve fine refinement result for certain good guidances but is not good at correcting the largely deviated guidances. The inference time is also shown in this table, which demonstrates the efficiency of our method.

\textbf{ALP:} 
Since DeepMANTA only provides their results evaluated in Average Localization Precision (ALP) metric \cite{deepmanta}, we also preform results comparison in this metric. As shown in Table.\ref{tab:loc}, our method is outstanding among current state of the art works, except that 3DOP outperforms us in this metric. Since $\text{ALP}$ focus only on the location accuracy and the size and rotation is not taken into consideration, its ability of reflecting the true quality of the 3D box may be not as good as 3D overlap. 

\begin{table}[!ht]
\begin{center}
\linespread{1.2}
\setlength{\tabcolsep}{1mm}
\footnotesize
\begin{tabular}{| c | c || c@{/}c | c@{/}c | c@{/}c |}
\hline
\multirow{2}{*}{Method} & \multirow{2}{*}{Extra} & \multicolumn{6}{c|}{$\text{ALP} _{1m}$} \\
\cline{3-8}
& & \multicolumn{2}{c|}{Easy} & \multicolumn{2}{c|}{Moderate} & \multicolumn{2}{c|}{Hard}  \\
\hline
3DVP \cite{3dvp} & None & 45.61 & - & 34.28 & - & 27.72 & - \\
Deep3DBox \cite{deep3dbox} & None & 35.71 & - & 25.35 & - & 23.03 & - \\
Mono3D \cite{mono3d} & Mask & - & 48.31 & - & 38.98 & - & 34.25 \\
DeepMANTA \cite{deepmanta} & CAD & 70.90 & 65.71 & 58.05 & 53.79 & 49.00 & 47.21 \\
3DOP \cite{3dop} & Stereo & - & \textbf{81.97} & - & \textbf{68.15} & - & \textbf{59.85} \\
\hline
Ours & None & \textbf{71.09} & 66.23 & \textbf{63.77} & 58.01 & 50.97 & 47.43 \\
Ours (scls) & None & 67.87 & 62.56 & 60.66 & 53.85 & \textbf{53.53} & 49.54 \\
\hline
\end{tabular}
\vspace{0.5mm}
\caption{3D detection for car category evaluated using the metric of $ALP$. Results on the two validation sets $val_1$ / $val_2$. ``Extra'' means the extra data or label used in training.}
\label{tab:loc}
\end{center}
\end{table}

\textbf{Results on Test Set:} 
Among all published monocular 3D detection works, only MF3D \cite{mfusion} shows the results evaluated on the official test set. The comparison between their results and ours is shown in Table.\ref{tab:test}.

We only submit once so there is no trick of hyper-parameter search. But even so, our method outperforms the other work. Note that both the results of MF3D and ours on test set have a gap compared with those on validation set (Table.\ref{tab:3d}). And this is most probably caused by the gap of data distribution between training and testing set, since KITTI training set is really small.

\begin{table}[!ht]
\begin{center}
\footnotesize
\begin{tabular}{| c | c | c | c |}
\hline
\multirow{2}{*}{Method} & \multicolumn{3}{c|}{AP$_{3D}$(IoU=0.7)} \\
\cline{2-4}
 & Easy & Moderate & Hard  \\
\hline
MF3D \cite{mfusion} & 7.08 & 5.18 & 4.68 \\
GS3D (Ours) & \textbf{7.69} & \textbf{6.29} & \textbf{6.16} \\
\hline
\end{tabular}
\vspace{0.5mm}
\caption{Our 3D detection results on official test set.}
\label{tab:test}
\end{center}
\end{table}

\subsection{Qualitative Results}
Fig.~\ref{fig:result} shows some qualitative results of our approach.
Our method can handle different scenes. It is robust to object in different distances from the camera. And when the scene is crowded, our method still performs well in most cases.
The red box in the two images in the last row shows a typical failure cases of our work. In the left image, the location of the box (in red) of the car on the bottom right corner has an obvious deviation from the true car. In the right image, the red dashed box is mistaken for negative box by our model. Our approach is not good at handling the objects on the boundary of the image (usually with occlusion or truncation). Further efforts is in need to solve this problem.

\section{Conclusions}
In this paper, we have proposed a monocular 3D object detection framework for autonomous driving. We utilize the mature 2D detection technology and projection knowledge to efficiently generate basic 3D box called guidance. Based on the guidance, further refinement is performed to achieve high accuracy. We take advantage of potential 3D structure information in surface feature that eliminate the representation ambiguity brought by only using 2D bounding box. And we reformulate the hard residual regression problem into classification, which is easier to be well-trained. And we use a quality aware loss to enhance the discriminative ability of model. Experiment shows that our framework achieves new state-of-the-art as a method using single RGB image without any extra data or label for training.

{\small
\bibliographystyle{ieee}
\bibliography{mybib}
}

\end{document}